\documentclass{article}

\usepackage{microtype}
\usepackage{graphicx}
\usepackage{subcaption}
\usepackage{booktabs} % for professional tables
\usepackage{multirow}
\usepackage[table]{xcolor}

\newcommand*{\thead}[1]{\multicolumn{1}{c}{\bfseries\begin{tabular}[c]{@{}c@{}}#1\end{tabular}}}

\usepackage{hyperref}

\usepackage[preprint]{icml2026}
\usepackage{amsmath}
\usepackage{amssymb}
\usepackage{mathtools}
\usepackage{amsthm}
\usepackage{titletoc}
\usepackage{bbm}
\usepackage[capitalize,noabbrev]{cleveref}

\theoremstyle{plain}

\theoremstyle{definition}

\theoremstyle{remark}

\usepackage[textsize=tiny]{todonotes}

\icmltitlerunning{RSAgent: Learning to Reason and Act for Text-Guided Segmentation via Multi-Turn Tool Invocations}

\begin{document}

\twocolumn[
  \icmltitle{RSAgent: Learning to Reason and Act for Text-Guided Segmentation 
   \\via Multi-Turn Tool Invocations}



  \icmlsetsymbol{equal}{†}
  \icmlsetsymbol{correspond}{*}

  \begin{icmlauthorlist}
    \icmlauthor{Xingqi He}{equal,ai}
    \icmlauthor{Yujie Zhang}{equal,fdu,sii}
    \icmlauthor{Shuyong Gao}{correspond,ai}
    \icmlauthor{Wenjie Li}{sii,sjtu}
    \icmlauthor{Lingyi Hong}{ai}
    \icmlauthor{Mingxi Chen}{ai}
    \icmlauthor{Kaixun Jiang}{robot}
    \icmlauthor{Jiyuan Fu}{ai}
    \icmlauthor{Wenqiang Zhang}{correspond,ai,robot}
  \end{icmlauthorlist}

  \icmlaffiliation{ai}{Shanghai Key Lab of Intelligent Information Processing, College of Computer Science and Artificial Intelligence, Fudan University}
  \icmlaffiliation{robot}{College of Intelligent Robotics and Advanced Manufacturing, Fudan University}
  \icmlaffiliation{fdu}{College of Computer Science and Artificial Intelligence, Fudan University}
  \icmlaffiliation{sii}{Shanghai Innovation Institute}
  \icmlaffiliation{sjtu}{College of Health Science and Technology, Shanghai Jiao Tong University School of Medicine}
  
  \icmlcorrespondingauthor{Shuyong Gao}{shuyongg@andrew.cmu.edu}
  \icmlcorrespondingauthor{Wenqiang Zhang}{wqzhang@fudan.edu.cn}

  \icmlkeywords{Machine Learning, ICML}

  \vskip 0.3in
]




\printAffiliationsAndNotice{}  % no special notice (required even if empty)


\begin{abstract}
Text-guided object segmentation requires both cross-modal reasoning and pixel grounding abilities. Most recent methods treat text-guided segmentation as one-shot grounding, where the model predicts pixel prompts in a single forward pass to drive an external segmentor, which limits verification, refocusing and refinement when initial localization is wrong.
To address this limitation, we propose \textbf{RSAgent}, an agentic Multimodal Large Language Model (MLLM) which interleaves \textbf{reasoning and action} for segmentation via multi-turn tool invocations. RSAgent queries a segmentation toolbox, observes visual feedback, and revises its spatial hypothesis using historical observations to re-localize targets and iteratively refine masks. We further build a data pipeline to synthesize multi-turn reasoning segmentation trajectories, and train RSAgent with a two-stage framework: cold-start supervised fine-tuning followed by agentic reinforcement learning with fine-grained, task-specific rewards. 
Extensive experiments show that RSAgent achieves a zero-shot performance of 66.5\% gIoU on ReasonSeg test, improving over Seg-Zero-7B by 9\%, and reaches 81.5\% cIoU on RefCOCOg, demonstrating state-of-the-art performance on both in-domain and out-of-domain benchmarks.

\end{abstract}

\section{Introduction}
\label{intro}
Text-guided segmentation takes an image and a natural language description of the task object as input and returns a fine-grained segmentation mask, covering both referring expression segmentation (RES) \cite{RefCOCO,RefCOCOg} and reasoning segmentation \cite{lisa_2024_Lai}, where the input text may be explicit expressions or implicit reasoning queries. Unlike closed-set semantic segmentation, which is restricted to a predefined label space, this setting must generalize to open-vocabulary concepts and therefore requires tight coupling between linguistic understanding and visual localization. This formulation introduces several key challenges: (i) grounding the task object to the correct region under large appearance variation, (ii) performing compositional or relational reasoning across modalities, and (iii) translating high-level linguistic cues into precise pixel boundaries. These requirements make text-guided segmentation particularly attractive for interactive perception and embodied applications (e.g., robotics). 

\begin{figure}
    \centering
    \includegraphics[width=1\linewidth]{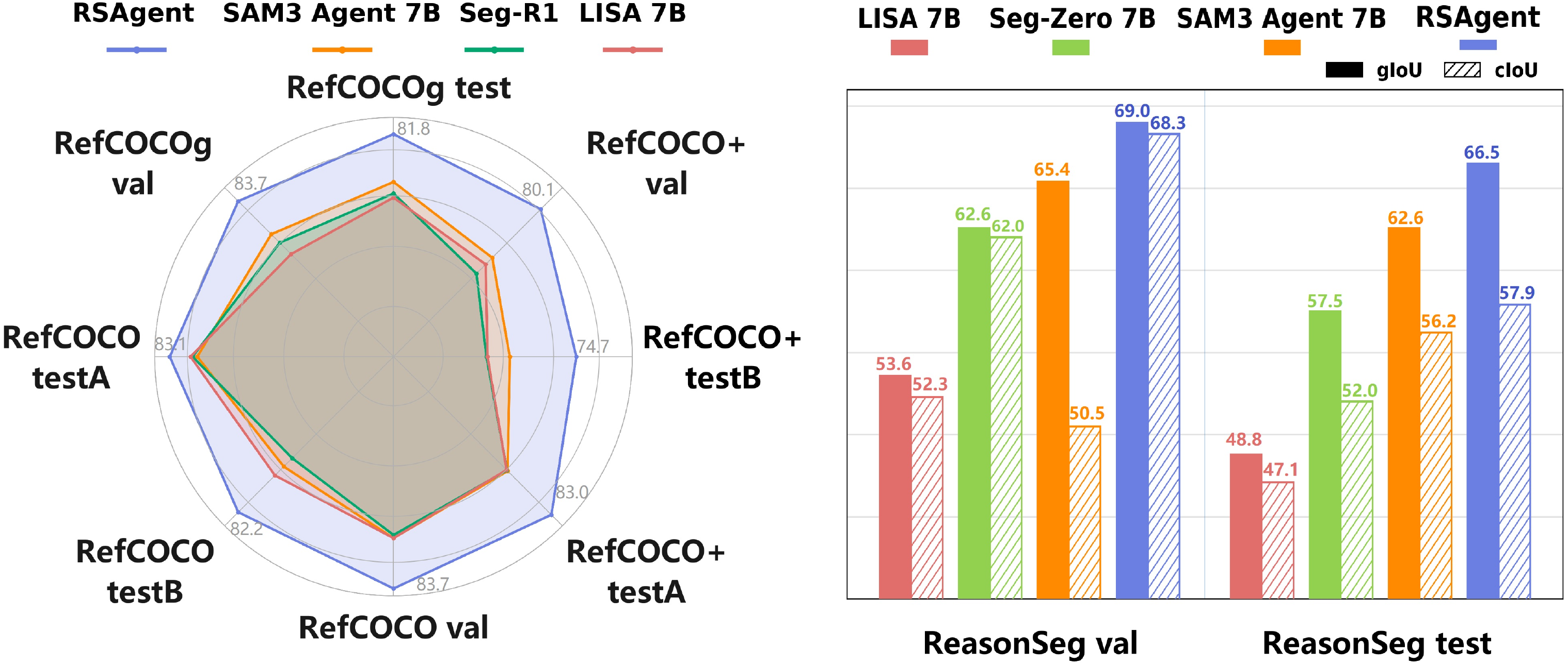}
    \caption{Performance on text-guided object segmentation benchmarks. RSAgent achieves state-of-the-art on both RES and ReasonSeg benchmarks.}
    \vspace{-8pt}
    \label{fig:table}
\end{figure}

\begin{figure*}[t]
    \centering
    \includegraphics[width=1\linewidth]{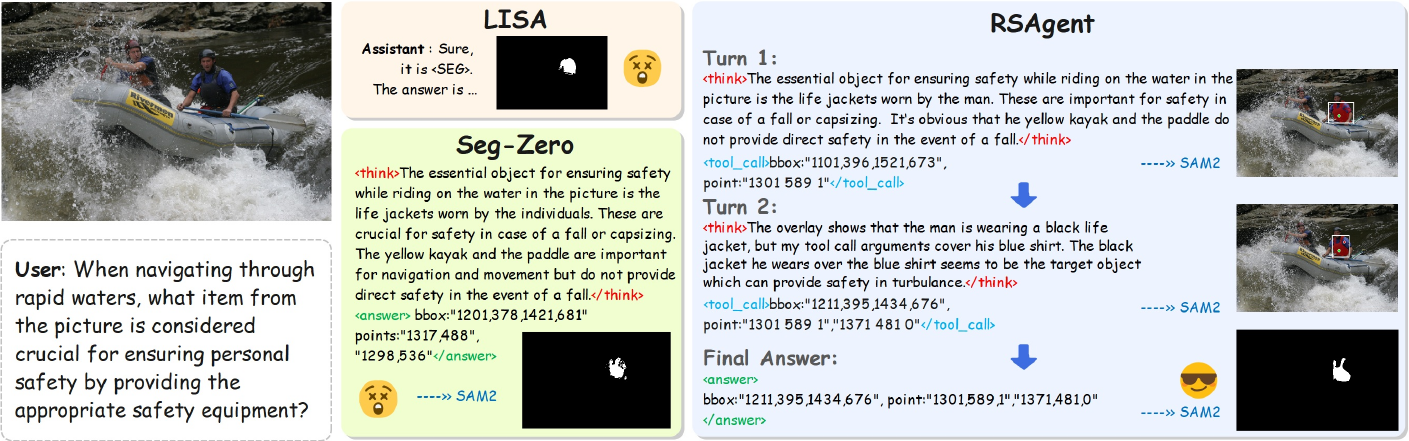}
    \caption{Comparing to LISA's direct segmentation and Seg-Zero's single forward pass of thinking and segmentation, RSAgent operates by iteratively proposing and updating spatial prompts and invoking external visual tools to iteratively refine the final mask. }
    \vspace{-3pt}
    \label{fig:examples}
\end{figure*}

Multimodal Large Language Models (MLLMs) have become a central component in the latest paradigms for this task because of their integrated reasoning and multimodal perception abilities \cite{pixellm2024,onetoken2024bai,glamm2024,mlcdseg2024,hyperseg2024,xie2025rice,wang2025xsam}. Early attempts such as LISA \cite{lisa_2024_Lai}, have explored the use of MLLMs to enhance reasoning segmentation capabilities via predefined semantic tokens (e.g., $<$SEG$>$ token), bridging the gap between MLLMs and segmentation models by supervised fine-tuning (SFT). Although SFT approaches can effectively inject MLLM reasoning into segmentation pipelines, they often face issues such as eroding general reasoning competence and weak robustness under distribution shifts. Reinforcement learning (RL) style approaches, meanwhile, optimize the model with reward signals for generating pixel prompts (e.g., boxes and points) and then feed them into Segment Anything Model (SAM) \cite{SAM2023kirillov} to produce the final prediction. The reward design evolves from pixel prompt's accuracy \cite{segzero} to the predicted mask's IoU with ground truth (GT) mask \cite{samr1,segr1}, which are effective in enhancing models' reasoning ability. However, these RL structures face two major limitations: (i) single prompt generation and directly obtaining the final reward not only completely forgoes SAM's inherent capability for iterative refinement, but also prevents the model from revisiting visual information it initially overlooked, particularly those seemingly irrelevant and ambiguous regions to initiate new segmentation attempts, and (ii) the reward function may over-emphasize coarse overlap or prompt-level accuracy, while providing limited supervision on reasoning process, which probably results in unexpected masks.

Therefore, to address these limitations, we present RSAgent, an agentic MLLM that performs interleaved cross-modal reasoning and action for segmentation. Specially, RSAgent reformulates text-guided segmentation as an interactive problem: as shown in Figure~\ref{fig:examples}, given an image and a problem, the MLLM does not output mask tokens, nor rely on single-pass prompting, but instead produces coherent textual reasoning and decides tool actions to query an external segmentation toolbox, receiving visual feedback after each turn and using it for visual reflection. This iterative loop allows RSAgent to re-localize when early spatial hypotheses are wrong or refine the mask according to visual reflection. To make such agentic behavior learnable, we first introduce a data pipeline to generate high quality multi-turn reasoning segmentation trajectories. Then we construct a high quality dataset including 5K complete reasoning trajectories for cold-start SFT and 2K image-problem pairs as part of RL training dataset. Further we develop a two-stage training framework: a cold-start SFT phase that utilizes constructed multi-turn tool trajectories, followed by RL that optimizes long-horizon decisions with rewards that revisit while encouraging continuous mask improvement. As shown in Figure~\ref{fig:table}, RSAgent achieves state-of-the-art performance on both RES and ReasonSeg benchmarks, demonstrating that our approach is a stronger and more robust pathway for text-guided segmentation.

Our contributions can be listed as follows:
\begin{itemize}
\item We propose RSAgent, a novel agentic framework which enables the MLLM to query a segmentation toolbox, observe visual feedback, and revise its spatial hypothesis using historical observations to re-localize targets and iteratively refine masks.
\item We present a multi-turn reasoning trajectory generation pipeline, which helps to build the dataset for training. What's more, to enhance the agent's cross-modal reasoning ability, we introduce a two-stage training strategy: cold-start SFT and RL with fine-grained rewards.
\item We conduct extensive in-domain and out-of-domain experiments to evaluate RSAgent. Results show that RSAgent achieves 81.5\% average cIoU on RefCOCOg, 66.5\% gIoU on ReasonSeg test, establishing new state-of-the-art performance for text-guided segmentation.
\end{itemize}

\section{Related Work}
\label{related work}

\paragraph{Reasoning in Large Language Models.}
Recent LLMs have exhibited strong gains in deliberate reasoning: scaling inference time computation can reliably improve performance on challenging problems \cite{openaio12024learning_to_reason}, and process-aware supervision or verification further strengthens step-by-step solution quality \cite{ProcessOutcomeFeedback2022,LetsVerifyStepByStep2024,MathShepherd2024}. Beyond purely text-only reasoning, reason-and-act paradigms such as ReAct \cite{react2023yao} interleave natural language rationales with actions, enabling models to revise intermediate hypotheses using external observations rather than relying solely on token histories. RL based post-training can further amplify these behaviors: DeepSeek-R1 \cite{deepseekr1_nature_guo2025} demonstrates that RL can incentivize long-form reasoning patterns including self-reflection and verification. However, most prior work primarily targets high-level correctness and coarse-grained multimodal understanding \cite{pixelreasoner2025wang,hong2025deepeyesv2}, leaving dense, pixel-level reasoning tasks such as segmentation relatively underexplored. In contrast, RSAgent extends the “reason-and-act” paradigm to segmentation: it preserves coherent textual reasoning while interacting with a decoupled segmentation toolbox, enabling visual reflection for iterative re-localization and refinement.

\paragraph{MLLMs for Text-Guided Segmentation.}
Recent studies endow MLLMs with pixel-level segmentation capability from natural language prompts, typically by augmenting an MLLM with a mask generation module, e.g., a promptable segmentor such as SAM or a lightweight mask decoder. Early token-based approaches introduce special segmentation tokens (e.g., \texttt{<SEG>}) and an embedding-as-mask interface that maps language-model representations to dense masks \cite{lisa_2024_Lai,onetoken2024bai,mlcdseg2024,wang2025xsam}. Building on this line, MLLM-centric systems further move toward unified pixel grounding under conversational and multi-target settings, such as PixelLM \cite{pixellm2024} and GLaMM \cite{glamm2024}. GSVA \cite{gsva2024} extends referring segmentation to the generalized setting by supporting multiple targets and explicitly rejecting empty referents. Complementary to SFT style training, several recent methods emphasize RL for learning a decoupled policy that outputs spatial prompts to guide an external segmentor: Seg-Zero \cite{segzero} produces position prompts with RL, SAM-R1 \cite{samr1} leverages mask reward feedback for fine-grained alignment, and VisionReasoner \cite{visionreasoner2025liu} unifies multiple perception tasks in an RL-based framework. Besides, POPEN \cite{popen2025zhu} explores preference-based optimization and ensembling to improve segmentation quality and reduce hallucinations.

\begin{figure*}
    \centering
    \includegraphics[width=1\linewidth]{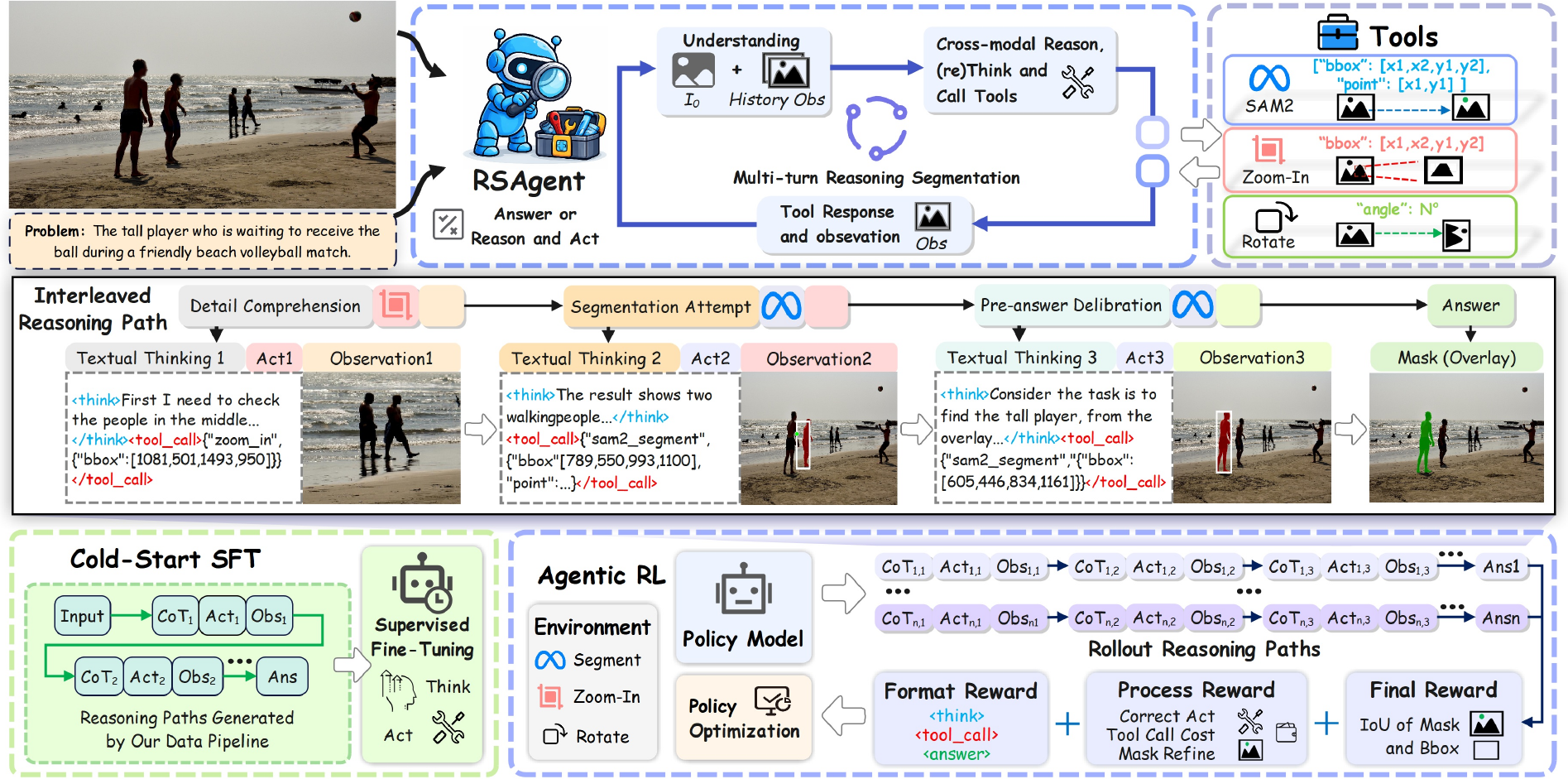}
    \caption{\textbf{Overview of RSAgent.}  Given the original image and problem, the agent interacts with an external visual toolbox over multiple rounds, incrementally gathers visual evidence, refines candidate masks, and eventually committs to a final prediction. RSAgent first embraces cold-start SFT to get accustomed to reasoning and multi-turn tool invocating operations via the cold-start data generated by our data pipeline, then gets optimized by RL with fine-grained rewards.}
    \label{fig:arch}
\end{figure*}

\section{Method}
\label{sec:method}

In this section, we present RSAgent, the data pipeline and the training strategies that enable effective multi-turn tool invocations for text-guided segmentation. We first formulate text-guided segmentation as an episodic decision making problem in pixel space (Section~\ref{sec:problem formulation}).  Then we provide an overview of RSAgent (Section~\ref{sec:overview}). Next, we describe our data pipeline and cold-start SFT that bootstrap tool-use behaviors and align the base model to the interactive setting (Section~\ref{sec:coldstart}). Finally, we introduce the RL stage (Section~\ref{sec:rl}), which further improves multi-round refinement and encourages efficient tool invocation.

\subsection{Problem Formulation}
\label{sec:problem formulation}
Unlike prior approaches that predict a mask in a single forward pass of the model \cite{segzero,segr1}, we cast text-guided segmentation as an episodic decision making problem in pixel space. As illustrated in Figure~\ref{fig:arch}, the agent interacts with an external visual toolbox over multiple rounds, incrementally gathers visual evidence, refines candidate masks, and eventually commits to a final prediction. Formally, each example consists of an image $I$, a natural language problem $Q$ which could be either an explicit referring expression or a complex description, and, during training only, a GT mask $M_{\mathrm{gt}}$.

\subsection{Overview of RSAgent}
\label{sec:overview}
\paragraph{Agent tool interaction.}
We formalize RSAgent as a vision-language policy $\pi_\theta$ operating in a finite-horizon Markov Decision Process (MDP). At step $t$, the agent receives an observation $O_t = (V_t, Q, C_t)$, where $V_t$ denotes the set of visual views available to the agent (including the original image and a history pool of overlays summarizing previous observations), and $C_t$ is the text context composed of the system prompt, the user prompts, and the accumulated assistant's feedback up to step $t$. For notational convenience, we write the reasoning trajectory up to step $T$ as:

\begin{equation}
\label{eq:episode}
P_{1:T} = \{(O_{t-1},r_t, a_t, o_t)\}_{t=1}^{T},
\end{equation}
where $r_t$ denotes the model’s intermediate reasoning tokens, $a_t$ denotes a set of parameterized tool calls and $o_t$ means the observation of this turn, which is concated to $O_t$.
In the whole $P_{1:T}$ , RSAgent alternates between textual reasoning, tool usage, and inspection of updated visual context, gradually refining its belief about the target object before committing to a final mask prediction.

\paragraph{Action space and tools.}
Conditioned on $O_{t-1}$, the policy $\pi_\theta$ generates an interleaved sequence of reasoning and action tokens, which may contain (i) updated reasoning $r_t$ together with one or more structured tool invocations $a_t$, or (ii) a terminating answer of spatial prompts describing the final segmentation. The tools exposed to the agent form a compact but expressive action space: 
(i) view-manipulation operations that zoom in or rotate the image to adjust the field of view, and 
(ii) the major and frequently used segmentation operation that receives spatial prompts into candidate masks via a frozen SAM2 \cite{ravi2025sam2}. The episode terminates when the agent emits a dedicated answer or when a maximum interaction budget is reached.

\subsection{Data Pipeline and Cold-Start Supervised Fine-tuning}
\label{sec:coldstart}

\paragraph{Problem-centric data collection.}
To synthesize the multi-turn reasoning trajectories, constructing the textual promblems for images is the primary task. We first collect image–mask pairs $(I, M_{\mathrm{gt}})$ and associated annotations from SA-1B dataset \cite{SAM2023kirillov}. Then we construct a natural language problem $Q$ describing the target object. Concretely, we employ strong proprietary vision-language models (practically, we used Gemini2.5-Pro \cite{Gemini25Pro} and OpenAI-o3 \cite{o3}) to generate problems in the style of reasoning segmentation according to $(I, M_{\mathrm{gt}})$. The details are provided in Appendix.~\ref{append:data}.

\begin{figure}
    \centering
    \includegraphics[width=0.9\linewidth]{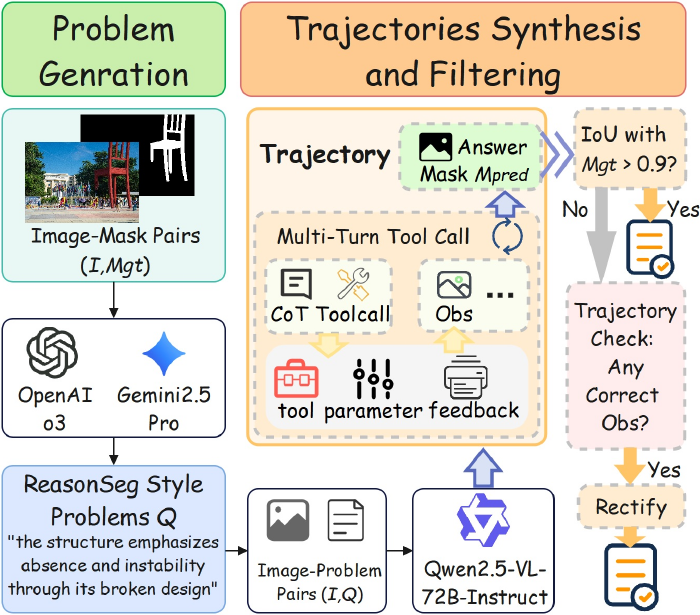}
    \caption{The multi-turn reasoning segmentation data pipeline, including problem generation, trajectory synthesis and data filtering.}
    \label{fig:datapipeline}
\end{figure}

\paragraph{Multi-turn tool interaction for trajectory synthesis.}
After problem generation, we synthesize full multi-round trajectories with the seed pair $(I, Q)$ by letting a model with the ability of pixel understanding interact with the same tool environment as RSAgent and generate multi-turn interleaved reasoning trajectories. Concretely, we place Qwen2.5-VL-72B-Instruct \cite{bai2025qwen25vl} in our visual tool environment, which exposes the view manipulation and segmentation tools described in Section~\ref{sec:overview}. The resulting interaction yields a full pixel-space reasoning trajectory with multi-turn tool calls and corresponding visual states. The overall pipeline is illustrated in Figure~\ref{fig:datapipeline}.

\paragraph{Trajectory filtering.}
To ensure that the synthetic trajectories provide reliable supervision, we select the cold-start data with two hard principles: (i) the IoU of the final predicted mask and $M_{\mathrm{gt}}$ can't be lower than 0.9, and (ii) the number of reasoning turns should not surpass 8. The former principle not only validates the accuracy of the problem, but also demonstrates the multi-turn interleaved reasoning process's correctness. For the latter, although some trajectory's final answer is correct, the actual reasoning process may be unreliable due to overly long context, compounding errors or hallucinated rationales.

In addition to the high quality trajectories selected under the above criteria, we also filter a supplemental set of trajectories that exhibit a modest number of tool invocations and a low final mask IoU, for which a correct mask is produced by segmentation tool at some intermediate reasoning step. After manually verifying the correctness of the corresponding problem $Q$, we revise the terminal supervision to align with this correct intermediate mask and discard the most misleading portions of the erroneous reasoning trace, and then add these curated trajectories to our dataset.

\paragraph{Cold-start SFT.}
Through our data pipeline, we obtain a dataset $D_{\mathrm{sft}}$ of approximately 5K examples, each containing both coherent reasoning trajectories and corrective refinement traces for cold-start SFT. The training objective is to minimize the average negative log-likelihood over all reasoning and action tokens in the cold-start trajectories. 
% Concretely, given a full trajectory $P_N$ from $D_{\mathrm{sft}}$, we define a token mask $m_t\in\{0,1\}$
% that selects reasoning and tool-call tokens, and optimize:
Concretely, given a trajectory $P$ with token sequence $y_{1:|P|}$, we define a token mask $m_t\in\{0,1\}$
to select reasoning and tool-call tokens, and optimize:
\begin{equation}
\label{eq:sft}
\begin{split}
\mathcal{L}_{\mathrm{SFT}}(\theta)
&=
\mathbb{E}_{P\sim \mathcal{D}_{\mathrm{sft}}}
\\
\Biggl[
&-\frac{1}{\sum_{t=1}^{|P|} m_t}\sum_{t=1}^{|P|}
m_t \log \pi_\theta\!\left(y_t \mid y_{<t}\right)
\Biggr].
\end{split}
\end{equation}

This objective teaches the model to produce effective trajectories with robust
step-by-step reasoning patterns, thereby endowing RSAgent with stronger pixel-level reasoning
and self-correction capabilities and providing a solid foundation for subsequent RL.

\subsection{Agentic Tool Reinforcement Learning}
\label{sec:rl}
We further optimize RSAgent with RL under a carefully designed reward scheme, enabling the agent to adaptively discover effective tool-use strategies for courageous attempt and iterative mask refinement. 

\subsubsection{RL Dataset}
The essence of RL training is to let the policy learn how to select actions that yield higher returns through interaction with the environment. Consequently, unlike the cold-start SFT stage, the RL dataset does not need to contain pre-generated multi-round trajectories: it is sufficient to provide the policy with the image $I$ and its associated problem $Q$ as input, together with the GT mask $M_{\mathrm{gt}}$ used solely for reward computation.

We construct approximately 2K RL examples using the same data curation procedure as in Section~\ref{sec:coldstart}, and additionally sample 8K instances from the RefCOCOg training split to form the overall RL dataset $D_{\mathrm{rl}}$.

\subsubsection{Reward Design}
Unlike prior approaches \cite{segzero,samr1} that rely solely on outcome-based rewards for models that predict a segmentation mask in a single forward pass, we introduce a fine-grained reward design that jointly accounts for both the final result and the intermediate decision process, enabling the policy to acquire more efficient and effective decision strategies. Specifically, our reward function consists of the following components:
\begin{itemize}
  \item \textbf{Final-answer reward} $R_{\mathrm{final}}$:
  encourages high quality final masks via both mask IoU
  and bounding box IoU with the GT.

  \item \textbf{Format reward} $R_{\mathrm{format}}$:
  rewards syntactically valid, schema-compliant \texttt{<answer>} and
  \texttt{<tool\_call>} blocks, penalizes unparsable outputs and invalid tool invocations.

  \item \textbf{Process reward} $R_{\mathrm{process}}$:
  step-wise shaping based on IoU improvement, including the
  best-so-far IoU in the reasoning process, tool-dependent call costs,
  and point-level sparsity or novelty for segmentation prompts.
  
\end{itemize}
The unified reward function is formulated as:
\begin{equation}
\label{eq:Rtotal}
R_{\mathrm{total}} = \alpha \cdot R_{\mathrm{final}} + \beta \cdot R_{\mathrm{process}} + \gamma \cdot R_{\mathrm{format}},
\end{equation}
where $\alpha$, $\beta$, $\gamma$ controls the relative strength of dense shaping versus outcome quality. The details of each reward are provided in Appendix ~\ref{append:reward}. 

\subsubsection{Training Objective}
\label{sec:optimization}

Based on the rollout formulation in Section~\ref{sec:overview} and rewards defined above, we optimize the policy with
Group Relative Policy Optimization (GRPO) \cite{GRPOdeepseekmath2024shao} (without an explicit KL penalty term \cite{OpenReasonerZero2025} as commonly used in RLHF)
on dataset $\mathcal{D}_{\mathrm{rl}}$. Let $y_{i,1:T_i}$ denote the concatenation of all non-observation tokens (reasoning tokens and tool-call tokens) generated in rollout $P_i$. The GRPO objective is:
\begin{equation}
\begin{split}
\mathcal{L}_{\mathrm{RL}}(\theta)
=
\mathbb{E}_{\substack{(I,Q,M_{\mathrm{gt}})\sim \mathcal{D}_{\mathrm{rl}}\\ \{P_i\}_{i=1}^{G}\sim \pi_{\theta_{\mathrm{old}}}(\cdot \mid I,Q)}} 
\Big(
-\frac{1}{G}\sum_{i=1}^{G}\frac{1}{T_i}\sum_{n=1}^{T_i}\\
\min\!\big(\rho_{i,n} A_i,\; \operatorname{clip}(\rho_{i,n},1-\epsilon,1+\epsilon)\,A_i\big)
\Big),
\end{split}
\end{equation}

where $\rho_{i,n}$ denotes the importance ratio:
\begin{equation}
\label{eq:grpo_ratio}
\rho_{i,n}=
\frac{\pi_{\theta}(y_{i,n}\mid y_{i,<n}, I,Q)}
{\pi_{\theta_{\mathrm{old}}}(y_{i,n}\mid y_{i,<n}, I,Q)} .
\end{equation}

Here, $G$ is the number of sampled rollout paths; $T_i$ is the number of non-observation tokens in $P_i$; and $S_i$ is the trajectory-level return computed from our reward.
We compute a group-relative advantage $A_i$ by normalizing returns within each rollout group:
\begin{equation}
\label{eq:grpo_adv}
\begin{aligned}
A_i &= \frac{S_i-\mu_S}{\sigma_S}, \qquad
\mu_S = \frac{1}{G}\sum_{j=1}^{G} S_j,\\
\sigma_S &= \sqrt{\frac{1}{G}\sum_{j=1}^{G}(S_j-\mu_S)^2+\delta},
\end{aligned}
\end{equation}
where $\delta$ is a small constant for numerical stability. Through carefully designed RL training, the agent becomes capable of interpreting target descriptions and performing pixel-level reasoning via iterative tool invocations, thereby accurately segmenting the object in the image that best matches the textual problem.

\begin{table*}[t]
\centering
\small
\renewcommand{\arraystretch}{0.95}
\setlength{\tabcolsep}{4pt}

\caption[Comparison with previous state-of-the-art methods on RES benchmarks.]{
Comparison with previous state-of-the-art methods on RES benchmarks.
We report cIoU (\%) on RefCOCO, RefCOCO+~\protect\cite{RefCOCO} and RefCOCOg~\protect\cite{RefCOCOg}.
The overall best performances are shown in \textbf{bold}, while the second best performances are shown \underline{underlined}.}

\label{tab:refcoco}

\begin{tabular}{l |l |c |c |ccc |ccc |cc}
\toprule
\multicolumn{1}{c}{\multirow[c]{2}{*}[-0.2pt]{\textbf{Method}}} &
\multicolumn{2}{c}{\textbf{Model}} &
\multicolumn{1}{c}{\multirow[c]{2}{*}[-1.5pt]{\textbf{\shortstack[c]{RES\\\\Train}}}} &
\multicolumn{3}{c}{\textbf{RefCOCO}} &
\multicolumn{3}{c}{\textbf{RefCOCO+}} &
\multicolumn{2}{c}{\textbf{RefCOCOg}} \\
\cmidrule(lr){2-3}\cmidrule(lr){5-7}\cmidrule(lr){8-10}\cmidrule(lr){11-12}
& Version & Params & &
val & testA & testB &
val & testA & testB &
val & test \\
\midrule

\multicolumn{12}{c}{\textbf{\textit{Discriminative vision--language segmentors without LLM-based controllers.}}} \\
\midrule
LAVT      & --          & --   & Y & 72.7 & 75.8 & 68.8 & 62.1 & 68.4 & 55.1 & 61.2 & 62.1 \\
ReLA      & --          & --   & Y & 73.8 & 76.5 & 70.2 & 66.0 & 71.0 & 57.7 & 65.0 & 66.0 \\
DETRIS    & DETRIS-L    & --   & Y & 81.0 & 81.9 & 79.0 & 75.2 & 78.6 & 70.2 & 74.6 & 75.3 \\
UniLSeg   & UniLSeg-100 & --   & Y & 81.7 & 83.2 & 79.9 & 73.2 & 78.3 & 68.2 & 79.3 & \underline{80.5} \\
EVF-SAM   & Extra Data  & --   & Y & 82.4 & \underline{84.2} & 80.2 & 76.5 & 80.0 & 71.9 & 78.2 & 78.3 \\
\addlinespace[2pt]
\midrule

\multicolumn{12}{c}{\textbf{\textit{Large (multi-)modal models for single-pass segmentation without explicit reasoning traces.}}} \\
\midrule
LISA      & LLaVA1.5    & 7B   & Y & 74.9 & 79.1 & 72.3 & 65.1 & 70.8 & 58.1 & 67.9 & 70.6 \\
GSVA      & LLaVA1.5    & 13B  & Y & 79.2 & 81.7 & 77.1 & 70.3 & 73.8 & 63.6 & 75.7 & 77.0 \\
GLaMM     & Vicuna      & 7B   & Y & 79.5 & 83.2 & 76.9 & 72.6 & 78.7 & 64.6 & 74.2 & 74.9 \\
SAM4MLLM  & LLaVA1.6    & 8B   & Y & 79.8 & 82.7 & 74.7 & 74.6 & 80.0 & 67.2 & 75.5 & 76.4 \\
PixelLM   & LLaVA2      & 7B   & Y & 73.0 & 76.5 & 68.2 & 66.3 & 71.7 & 58.3 & 69.3 & 70.5 \\
RICE      & Qwen2.5     & 7B   & Y & \underline{83.5} & \textbf{85.3} & \underline{81.7} & \underline{79.4} & \underline{82.8} & \textbf{75.4} & \underline{79.8} & 80.4 \\
\addlinespace[2pt]
\midrule

\multicolumn{12}{c}{\textbf{\textit{Explicit chain-of-thought (CoT) or reasoning-chain guided segmentation.}}} \\
\midrule
Seg-Zero  & Qwen2.5     & 7B   & Y & --   & 80.3 & --   & --   & 76.2 & --   & 72.6 & --   \\
Seg-R1    & Qwen2.5     & 7B   & N & 74.3 & 78.7 & 67.6 & 62.6 & 70.9 & 57.9 & 71.0 & 71.4 \\
\textit{RSAgent-single}  & Qwen2.5     & 7B   & Y & 80.6   & 81.1 & 78.9   & 77.3   & 81.1 & 70.4   & 73.7 & 76.0  \\
\addlinespace[2pt]
\midrule

\multicolumn{12}{c}{\textbf{\textit{Multi-round tool-calling agents with iterative refinement.}}} \\
\midrule
SAM3 Agent & Qwen2.5     & 7B   & N & 53.4 & 58.4 & 48.0 & 46.3 & 52.2 & 40.8 & 54.5 & 55.1 \\
SAM3 Agent & Gemini2.5-Pro   & -- & N & 74.9 & 77.8 & 69.9 & 66.9 & 71.1 & 62.4 & 73.3 & 73.6 \\
\midrule
\rowcolor{gray!30}
\textit{RSAgent (ours)} & Qwen2.5-VL & 7B & Y &
\textbf{83.7} & 83.1 & \textbf{82.2} &
\textbf{80.1} & \textbf{83.0} & \underline{74.7} &
\textbf{81.3} & \textbf{81.8} \\
\bottomrule
\end{tabular}
\end{table*}

\section{Experiment}
\label{sec:experiment}
\subsection{Experimental Settings.}
\paragraph{Implementation Details.}
We implement our RSAgent based on Qwen2.5-VL-7B-Instruct \cite{bai2025qwen25vl}. SAM2-large \cite{ravi2025sam2} is used as the segmentation tool. In cold-start SFT stage, we optimize the model with a learning rate of 2 × $10^{-5}$ for 2 epochs and the batch size is 128. After the SFT stage, we implement the RL optimization on VERL \cite{HybridFlow2025} framework with batch size of 16 and sampling number of 4. The maximum context length is 32K and learning rate is 1 × $10^{-6}$. Moreover, the max number of tool call turns during training is 8. Whether in RL training or evaluation stage, direct answer will be given if the number of turns exceeds it.

\paragraph{Benchmarks and Datasets.}
We mainly conduct experiments on two benchmarks: RES benchmarks (RefCOCO series\cite{RefCOCO,RefCOCOg}, which includes RefCOCO, RefCOCO+ and RefCOCOg) and ReasonSeg\cite{lisa_2024_Lai} benchmarks. To ensure fair evaluation and comparison with previous work, we only randomly sampled 8K data from RefCOCOg train split as part of $D_{\mathrm{rl}}$ and keep ReasonSeg dataset out-of-domain.  The cold-start dataset $D_{\mathrm{sft}}$ and the rest of $D_{\mathrm{rl}}$ dataset are all generated with random samples from SA-1B by the data pipeline introduced in Section~\ref{sec:coldstart}.

\paragraph{Evaluation Metrics.}
Following previous work on text-guided segmentation, we adopt two commonly used evaluation metrics: generalized IoU (gIoU) and cumulative IoU (cIoU). Specifically, gIoU is computed as the average of per-image IoU, while cIoU is defined as the IoU of the cumulative predicted and GT masks across the entire dataset.

\paragraph{Baselines.}
We compare RSAgent with four families of representative methods on both RES and ReasonSeg benchmarks: (i) discriminative vision--language segmentors without LLM-based controllers, (ii) large (multi-)modal models for single-pass segmentation without explicit reasoning traces, (iii) Explicit chain-of-thought (CoT) or reasoning-chain guided segmentation, and (iv) multi-round tool-calling agents with iterative refinement.  What's more, we restrict RSAgent to output mask in single forward pass as \textit{RSAgent-single} as additional of our baselines. Most prior RES methods are fully supervised on the entire RES training sets, while several reasoning-segmentation models additionally leverage large-scale grounded conversation or reasoning-segmentation corpora. The detailed information of our baseline methods is introduced in Appendix \ref{append:baseline}.

% preamble:
% \usepackage{booktabs}
% \usepackage{multirow}

\begin{table*}[t]
\centering
\small
\renewcommand{\arraystretch}{0.95}
\setlength{\tabcolsep}{4pt}

\caption{Comparison with state-of-the-art methods on the ReasonSeg benchmark \cite{lisa_2024_Lai}.
We report generalized IoU (gIoU) and cumulative IoU (cIoU) in \%. The overall best performances are shown in \textbf{bold}, while the second best performances are shown \underline{underlined}.}
\label{tab:reasonseg}

\begin{tabular}{l |l |c |c |c |cc |cc |cc |cc}
\toprule
\multicolumn{1}{c}{\multirow[c]{2}{*}[-0.2pt]{\textbf{Method}}} &
\multicolumn{2}{c}{\textbf{Model}} &
\multicolumn{2}{c}{\textbf{Training}} &
\multicolumn{2}{c}{\textbf{Val}} &
\multicolumn{2}{c}{\textbf{Test}} &
\multicolumn{2}{c}{\textbf{Test (Short)}} &
\multicolumn{2}{c}{\textbf{Test (Long)}} \\
\cmidrule(lr){2-3}\cmidrule(lr){4-5}\cmidrule(lr){6-7}\cmidrule(lr){8-9}\cmidrule(lr){10-11}\cmidrule(lr){12-13}
& Version & Params & RES & ReasonSeg
& gIoU & cIoU & gIoU & cIoU & gIoU & cIoU & gIoU & cIoU \\
\midrule

\multicolumn{13}{c}{\textbf{\textit{Discriminative vision--language segmentors without LLM-based controllers.}}} \\
\midrule
SEEM         & --         & --   & N & N & 25.5 & 21.2 & 24.3 & 18.7 & 20.1 & 11.5 & 25.6 & 20.8 \\
Grounded-SAM & --         & --   & N & N & 26.0 & 14.5 & 21.3 & 16.4 & 17.8 & 10.8 & 22.4 & 18.6 \\
OVSeg        & --         & --   & N & N & 28.5 & 18.6 & 26.1 & 20.8 & 18.0 & 15.5 & 28.7 & 22.5 \\
\addlinespace[2pt]
\midrule

\multicolumn{13}{c}{\textbf{\textit{Large (multi-)modal models for single-pass segmentation without explicit reasoning traces.}}} \\
\midrule
GLaMM    & Vicuna   & 7B   & Y & N & 47.4 & 47.2 & --   & --   & --   & --   & --   & --   \\
SAM4MLLM & LLaVA1.6 & 8B   & Y & N & 58.4 & 60.4 & --   & --   & --   & --   & --   & --   \\
X-SAM    & Phi-3    & 3.8B & Y & Y & 56.6 & 32.9 & 57.8 & 41.0 & 47.7 & 48.1 & 56.0 & 40.8 \\
HyperSeg & Phi-2    & 3B   & Y & Y & 59.2 & 56.7 & --   & --   & --   & --   & --   & --   \\
LISA     & LLaVA1.5 & 7B   & Y & N & 53.6 & 52.3 & 48.8 & 47.1 & 48.3 & 48.8 & 49.2 & 48.9 \\
LISA     & LLaVA1.5 & 7B   & Y & Y & 61.3 & 62.9 & 51.7 & 51.1 & 44.3 & 42.0 & 57.9 & 59.7 \\
\addlinespace[2pt]
\midrule

\multicolumn{13}{c}{\textbf{\textit{Explicit chain-of-thought (CoT) or reasoning-chain guided segmentation.}}} \\
\midrule
RSVP     & LLaVA1.6  & 7B  & N & N & 59.2 & 56.7 & 56.9 & 50.7 & 47.9 & 42.0 & 58.4 & 53.0 \\
RSVP     & GPT-4o    & --  & N & N & 64.7 & \underline{63.1} & 60.3 & \textbf{60.0} & 55.4 & \underline{50.4} & 61.9 & \textbf{62.5} \\
Seg-Zero & Qwen2.5-VL & 7B & Y & N & 62.6 & 62.0 & 57.5 & 52.0 & --   & --   & --   & --   \\
Seg-R1   & Qwen2.5-VL & 7B & N & N & 58.6 & 41.2 & 56.7 & 53.7 & --   & --   & --   & --   \\
SAM-R1   & Qwen2.5-VL & 7B & Y & N & 64.0 & 55.8 & 60.2 & 54.3 & --   & --   & --   & --   \\
\textit{RSAgent-single} & Qwen2.5-VL & 7B & Y & N & 58.2 & 53.1 & 56.1 & 52.6 & 54.9   & 47.7   & 62.1   & 56.3   \\
\addlinespace[2pt]
\midrule

\multicolumn{13}{c}{\textbf{\textit{Multi-round tool-calling agents with iterative refinement.}}} \\
\midrule
SAM3-Agent & Qwen2.5-VL & 7B & N & N & \underline{65.4} & 50.5 & \underline{62.6} & 56.2 & \underline{59.1} & 41.8 & \underline{63.7} & 57.8 \\
\midrule
\rowcolor{gray!30}
\textit{RSAgent (ours)} & Qwen2.5-VL & 7B & Y & N
& \textbf{69.0} & \textbf{68.3} & \textbf{66.5} & \underline{57.9} & \textbf{60.3} & \textbf{54.4} & \textbf{68.4} & \underline{60.1} \\
\bottomrule
\end{tabular}
\end{table*}

\subsection{Main Results}
\label{results}

\paragraph{Results on Referring Segmentation Benchmarks.}
As shown in Table~\ref{tab:refcoco}, RSAgent achieves competitive performance on RefCOCO and RefCOCO+, and particularly state-of-the-art results on RefCOCOg. Despite being trained on only a 8K subset of RefCOCOg, RSAgent matches or surpasses several MLLM-based baselines built on comparable or larger backbones like RICE \cite{xie2025rice}, substantially outperforms the CoT guided single-pass segmentation models such as Seg-Zero \cite{segzero} and greatly surpasses SAM3 Agent \cite{carion2025sam3} which only utilizes prompt engineering to regulate the model without training. What's more, RSAgent-single's performance achieves competitive performance and even outperforms the previous CoT guided single-pass segmentation models, demonstrating our method successfully enhanced the model's reasoning ability. As previous work tell that the GT annotations in RefCOCO(+/g) are not precise enough, which suggests that our RSAgent model should, in principle, achieve better performance than values in the table. We attribute these gains to two factors: (i) the multi-round tool invocation, which allows RSAgent to zoom in, segment and iteratively refine segmentation proposals conditioned on the action outcome, and (ii) the cold-start SFT which enable the model to reason-and-act and RL objective that explicitly rewards mask improvement and penalizes unnecessary tool calls.

\paragraph{Results on Reasoning Segmentation Benchmarks.}
As shown in Table~\ref{tab:reasonseg}, our RSAgent achieves state-of-the-art performance in ReasonSeg test and val in gIoU and competitive results in other splits. Quantitatively, RSAgent considerably outperforms earlier SAM2-based single-pass baselines such as \cite{segzero,samr1} by at least 5.0\% gIoU on val split, 6.3\% cIoU on val split, 6.3\% gIoU on test split and 3.6\% cIoU on test split, as it can handle long and compositional descriptions involving spatial relations or fine-grained attributes more robustly, demonstrating that explicit multi-round reasoning with a carefully designed toolbox and reward is highly effective for focusing on ambiguous regions before committing to a final mask. What's more, RSAgent-single's performance also outperforms the previous CoT guided single-pass segmentation models, which suggest that our method indeed enhanced the model's reasoning ability. Although SAM3 Agent drives SAM3 model to achieve strong performance on ReasonSeg through training-free multi-round tool invocations, RSAgent, equipped with efficient SFT and RL, can likewise operate in a multi-round tool-calling regime to control a SAM2-based toolbox and attain comparable, or even superior, evaluation results despite relying on the comparatively weaker SAM2 backbone, which reflects more holistic segmentation quality across objects and scenes.

\paragraph{Qualitative Analysis.}
Combining the above quantitative analysis and the qualitative exhibition of Figure~\ref{fig:examples}, existing segmentation approaches that rely on implicit tokens distort the native textual output space of MLLMs, thereby compromising their language capability and weakening semantic generalization, while single-pass of segmentation methods lack progressive interaction with the environment. More importantly, these methods suffer from a fundamental limitation in cross-modal reasoning, which prevents models from truly capturing fine-grained visual attributes. In contrast, RSAgent not only generates correct and coherent textual reasoning, but also performs visually grounded reflection through multi-turn tool invocations. This interleaved reasoning ability stems from our design to decouple reasoning from a segmentation-centric tool environment, which preserves the MLLM’s inherent language reasoning capacity while enabling adaptive exploration and refinement.

\subsection{Ablations}
\label{ablation}
\paragraph{Analysis of cold-start SFT and RL.}
We first verify whether cold-start SFT and RL are indispensable components of our training pipeline. As shown in Table~\ref{tab:ab_coldstart}, directly deploying an off-the-shelf Qwen2.5-VL-7B-Instruct model without any additional training in our framework yields poor performance on both RES and ReasonSeg benchmarks, and in some cases even underperforms the single forward pass baselines. We attribute this to a paradigm mismatch: the base model is primarily pretrained for QA-style generation and is not explicitly exposed to tool-use interactions, leading to significant misalignment when transferred to a novel tool invocation setting. In contrast, after cold-start SFT (or even RL alone), the model learns to invoke external tools to support interleaved reasoning segmentation. Moreover, the interactive nature of RL encourages more adaptive multi-turn behaviors, enabling the policy to autonomously explore, refine, and correct predictions across iterations.

\begin{table}[t]
\centering
\small
\setlength{\tabcolsep}{3pt}
\renewcommand{\arraystretch}{1.08}
\caption{Ablation on cold-start SFT and RL. Metrics are cIoU (\%).}
\label{tab:ab_coldstart}

\begin{tabular*}{\columnwidth}{@{\extracolsep{\fill}} l cc @{}}
\toprule
\thead{Setting} & \thead{ReasonSeg\\test} & \thead{RefCOCOg\\testA} \\
\midrule
single-pass baseline (1-pass)   & 43.9 & 64.9 \\
tool-agent (no training)        & 30.1 & 61.7 \\
\midrule
+ cold-start SFT only           & 55.4 & 73.9 \\
+ RL only                       & 54.3 & 77.3 \\
+ cold-start SFT + RL (full)    & \textbf{57.9} & \textbf{81.8} \\
\bottomrule
\end{tabular*}
\end{table}

\begin{figure}
    \centering
    \includegraphics[width=0.7\linewidth]{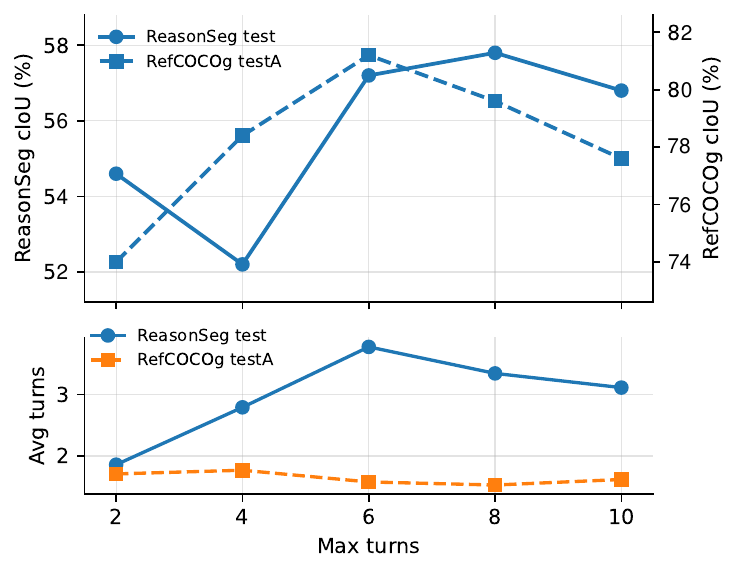}
    \caption{Effect of the maximum number of tool-invocation turns during training. Metrics are cIoU (\%).}
    \label{fig:ab_turn}
\end{figure}

\paragraph{Analysis of multi-turn tool invocation.}
Although multi-turn of decision in segmentation helps the model to rethink and provide accurate masks, overlong or unnecessary context may results in unexpected masks. As shown in Figure~\ref{fig:ab_turn}, the model's performance gradually improves as the max number of turns to call tools during training (max-turns) gets higher until 8 turns. What's more, during inference the average turns of successful predictions (when IoU $>$ 0.9) stops until max-turn reaches 6. For RefCOCOg benchmark which is less challenging compared with ReasonSeg, the tendency comes out the same, which indicates that appropriate max-turns during training do enhance the agent's cross-modal reasoning ability and incentivize its automony to invoke tools. 

% preamble:
% \usepackage{booktabs}
% \usepackage{multirow}

% \begin{table}[t]
% \centering
% \small
% \setlength{\tabcolsep}{1.8pt}      % 组内更紧（右侧4列更紧）
% \renewcommand{\arraystretch}{1.08}
% \caption{Effect of the maximum number of tool-invocation turns during training. Metrics are cIoU (\%). Avg turns are computed over successful predictions (IoU $>$ 0.9).}
% \label{tab:ab_turns}

% \begin{tabular}{@{} c cc @{\hspace{10pt}} cc @{}}
% \toprule
% \multirow[c]{2}{*}[-1.0pt]{\textbf{\mbox{Max-turns}}} &
% \multicolumn{2}{c}{\textbf{\mbox{ReasonSeg test}}} &
% \multicolumn{2}{c}{\textbf{\mbox{RefCOCOg testA}}} \\
% \cmidrule(lr){2-3}\cmidrule(lr){4-5}
% & \textbf{cIoU} & \textbf{Avg turns} & \textbf{cIoU} & \textbf{Avg turns} \\
% \midrule
% 2  & 56.3 & 1.86 & 80.0 & 1.71 \\
% 4  & 55.1 & 2.79 & 82.2 & \textbf{1.77} \\
% 6  & 57.6 & \textbf{3.77} & \textbf{83.6} & 1.58 \\
% 8  & \textbf{57.9} & 3.34 & 82.8 & 1.53 \\
% 10 & 57.4 & 3.11 & 81.8 & 1.62 \\
% \bottomrule
% \end{tabular}
% \end{table}

\paragraph{Analysis of reward design.}
We explore the impact of different reward functions during GRPO training. We study the model trained without $R_{\mathrm{final}}$, $R_{\mathrm{process}}$ and $R_{\mathrm{format}}$. Table~\ref{tab:ab_reward} shows the results that removing either component leads to consistent performance drops, demonstrating every reward's unique role in multi-turn reasoning segmentation ability construction.

\begin{table}[t]
\centering
\small
\setlength{\tabcolsep}{3pt}
\renewcommand{\arraystretch}{1.08}
\caption{Ablation on reward components for GRPO training. Metrics are cIoU (\%).}
\label{tab:ab_reward}

\begin{tabular*}{\columnwidth}{@{\extracolsep{\fill}} l cc @{}}
\toprule
\thead{Reward Variant} & \thead{ReasonSeg\\test} & \thead{RefCOCOg\\testA} \\
\midrule
Full ($R_{\mathrm{final}} + R_{\mathrm{process}} + R_{\mathrm{format}}$) & \textbf{57.9} & \textbf{81.8} \\
w/o $R_{\mathrm{final}}$                                                 & 48.3 & 68.1 \\
w/o $R_{\mathrm{process}}$                                               & 55.9 & 77.6 \\
w/o $R_{\mathrm{format}}$                                                & 54.7 & 80.3 \\
\bottomrule
\end{tabular*}
\end{table}

\begin{table}[t]
\centering
\small
\setlength{\tabcolsep}{3pt}
\renewcommand{\arraystretch}{1.08}
\caption{Ablation on segmentation tools. Metrics are cIoU (\%).}
\label{tab:ab_segmentor}

\begin{tabular*}{\columnwidth}{@{\extracolsep{\fill}} l cc @{}}
\toprule
\thead{Segmentation Model} & \thead{ReasonSeg\\test} & \thead{RefCOCOg\\testA} \\
\midrule
SAM2-hiera-large   & \textbf{57.9} & \textbf{81.8} \\
SAM2-hiera-base      & 56.7 & 80.9 \\
SAM2-hiera-tiny         & 53.1 & 77.2 \\
SAM-H                       & 53.3 & 74.6 \\
HQ-SAM-H    & 52.9 & 75.1 \\
\bottomrule
\end{tabular*}
\end{table}

\paragraph{Analysis of segmentation tool.}
We further investigate whether RSAgent’s gains mainly come from its multi-turn interleaved reason-and-act policy, rather than from a particular choice of the underlying segmentor. Specifically, we evaluate different versions of SAM and HQ-SAM \cite{HQSAM}. Table~\ref{tab:ab_segmentor} summarizes the results, showing that RSAgent consistently benefits from iterative interaction across tools, and can effectively exploit stronger segmentors while remaining robust when the tool is lightweight.

\section{Conclusion}
\label{conclusion}
We propose RSAgent, an agentic MLLM for text-guided segmentation that predicts masks via multi-turn reason-and-act with an external toolbox. Given an original image and textual problem, RSAgent operates by generating pixel prompts and invoking visual tools to focus on the correct region and iteratively refine the final mask. To enable the MLLM with the ability to correctly invoke appropriate tools and refine the mask through multi-turn of reasoning and actions, we first develop a data pipeline and generate dataset of 7K samples for training. Then we develop a two stage framework specialized for the multi-turn tool invocations. Qualitative results and quantitative evaluations demonstrate that RSAgent achieves state-of-the-art performance in both in-domain and out-of-domain benchmarks, revealing its interleaved reasoning segmentation ability to trial and refine via multi-turn tool invocations.

\bibliography{reference}
\bibliographystyle{icml2026}

%%%%%%%%%%%%%%%%%%%%%%%%%%%%%%%%%%%%%%%%%%%%%%%%%%%%%%%%%%%%%%%%%%%%%%%%%%%%%%%
%%%%%%%%%%%%%%%%%%%%%%%%%%%%%%%%%%%%%%%%%%%%%%%%%%%%%%%%%%%%%%%%%%%%%%%%%%%%%%%
% APPENDIX
%%%%%%%%%%%%%%%%%%%%%%%%%%%%%%%%%%%%%%%%%%%%%%%%%%%%%%%%%%%%%%%%%%%%%%%%%%%%%%%
%%%%%%%%%%%%%%%%%%%%%%%%%%%%%%%%%%%%%%%%%%%%%%%%%%%%%%%%%%%%%%%%%%%%%%%%%%%%%%%

\newpage
\appendix
\onecolumn

\section{Supplementary of RSAgent's Overview.}

\paragraph{Pixel-space memory and trajectory.}
When a segmentation tool is invoked, the environment selects one or more candidate masks and renders them back onto the original image as lightweight overlays (we choose the 384 × 384 as history pool's image size). These overlays are compressed into a fixed-size history pool and aggregated into $V_{t+1}$, so that subsequent decisions can attend jointly to the raw image and prior hypotheses. 
% In this way, a full episode is realized as a pixel-space reasoning trajectory:
% \begin{equation}
% \label{eq:episode}
% \tau = \{(O_t, a_t, r_t)\}_{t=1}^{T},
% \end{equation}
% in which the agent alternates between textual reasoning, tool usage, and inspection of updated visual context, gradually refining its belief about the target object before committing to a final mask prediction.

\paragraph{Termination and learning signal.}
The episode terminates when the agent emits a dedicated answer segment specifying which candidate masks constitute the final output (optionally accompanied by a textual justification), or when a maximum interaction budget is reached. At training time, the environment evaluates the final prediction $\hat{M}$ against $M_{\mathrm{gt}}$ and provides a segmentation-aware return that combines the terminal IoU-based score with dense, step-wise shaping terms reflecting mask improvement, sensible tool usage, and trajectory efficiency. This unified agent–tool formulation underlies both the cold-start supervised fine-tuning in Section~\ref{sec:coldstart} and the reinforcement learning stage in Section~\ref{sec:rl}.

\section{Details of Our Data Collection.}
\label{append:data}
To obtain supervision that is closely aligned with our multi-round RL environment, we construct a synthetic dataset of multi-step segmentation trajectories on top of a curated subset of SA-1B. The overall pipeline is illustrated in Figure~\ref{fig:datapipeline}.

\subsection{Problem-centric data collection.}
We build our cold-start data on top of the SA-1B dataset \cite{SAM2023kirillov}, a large-scale corpus of about 11M diverse images with 1.1B generated segmentation masks. We first select images from SA-1B in which at least one annotated instance occupies a moderate fraction of the image (roughly around $10\%$ of the image area). For each selected image, we choose a single target instance and treat its binary mask as the GT $M_{\mathrm{gt}}$, making image–mask pairs $(I, M_{\mathrm{gt}})$ and associated geometric annotations, which later serve as GT references for both mask evaluation and synthetic trajectory construction. Then we construct a natural language problem $Q$ describing the target object. Concretely, we employ strong proprietary vision–language models (practically, we used Gemini2.5-Pro \cite{Gemini25Pro} and OpenAI-o3 \cite{o3}) to generate problem in the style of reasoning segmentation according to $(I, M_{\mathrm{gt}})$.

\subsection{Multi-turn tool interaction for trajectory synthesis.}
After problem generation, we synthesize full multi-round trajectories with the seed pair $(I, Q)$ by letting a model with the ability of pixel-understanding interact with the same tool-augmented environment as RSAgent. Concretely, we place Qwen2.5VL-72B-Instruct \cite{bai2025qwen25vl} in our visual tool environment, which exposes the view manipulation and segmentation tools described in Section~\ref{sec:overview}. Our goal is to obtain multi-round segmentation trajectories that closely match the interaction protocol of RSAgent at test time: at the beginning of each episode, the Qwen-VL model receives the original image and the problem $Q$; at each subsequent step, it observes the updated visual context (the base image together with a small pool of overlays summarizing previous masks) and the accumulated textual conversation, and produces the next reasoning segment and tool invocations. The resulting interaction yields a full pixel-space reasoning trajectory with multi-turn tool calls and corresponding visual states, closely aligned with the Markov decision process used later for reinforcement learning. The overall pipeline is illustrated in Figure~\ref{fig:datapipeline}.

To ensure that the synthetic trajectories provide reliable supervision, we evaluate the final prediction against the $M_{\mathrm{gt}}$. We select the cold-start data with two hard principles: (i) the IoU of the final predicted mask's and $M_{\mathrm{gt}}$ can't be lower than 0.9, and (ii) the number of reasoning turns shouldn't surpass 8. The former principle not only validates the accuracy of the problem, but also demonstrates the multi-turn interleaved visual and textual reasoning process's rightness. Unlike some previous "think with images" work \cite{pixelreasoner2025wang,hong2025deepeyesv2,Ophiuchus2025,DeepEyes2025}, which make processive VQA tasks and don't provide "final-answer" form of observation in middle turns, every tool call of SAM2 will produce a final result in history pool. Therefore, although some trajectory's final answer is correct, the actual reasoning process may be unreliable due to overly long contexts and compounding errors / hallucinated rationales.

In complex scenes where the agent cannot backtrack or undo previous operations, any suboptimal decision at an intermediate step may propagate and compound, adversely affecting subsequent predictions and ultimately degrading the final segmentation quality. In addition to the high quality trajectories selected under the above criteria, we also incorporate a small supplemental set of trajectories that exhibit a modest number of tool invocations and a low final mask IoU, but for which a correct mask is produced by SAM2 at some intermediate reasoning step. After manually verifying the correctness of the corresponding problem $Q$, we revise the terminal supervision to align with this correct intermediate mask and discard the most misleading portions of the erroneous reasoning trace, and then add these curated trajectories to the cold-start dataset.

\section{Reward Design in RL Training}
\label{append:reward}

We formulate tool-augmented segmentation as a finite-horizon interactive episode.
At each step $t$, the policy either invokes a tool (zoom/rotate/segment) or outputs a final answer in the required JSON schema.
Our overall reward is decomposed into three components:
(i) \textbf{Final-answer reward} $R_{\mathrm{final}}$ for final mask quality,
(ii) \textbf{Format reward} $R_{\mathrm{format}}$ for valid \texttt{<think>} / \texttt{<tool\_call>} / \texttt{<answer>} blocks,
and (iii) \textbf{Process reward} $R_{\mathrm{process}}$ for step-wise shaping along the trajectory.

\subsection{Process reward $R_{\mathrm{process}}$.}
Following prior findings that step-level (process-based) supervision provides richer signals and eases long-horizon credit assignment compared to outcome-only feedback, we incorporate step-wise IoU-improvement shaping as a process reward to guide tool-use trajectories \cite{LetsVerifyStepByStep2024,ProcessOutcomeFeedback2022}.
Let $\mathrm{IoU}_t$ denote the foreground IoU (thresholded at $0.5$) between the current best predicted mask and the ground-truth mask after executing the $t$-th turn of tool call, and define $\Delta_t=\mathrm{IoU}_t-\mathrm{IoU}_{t-1}$.
We use step-wise shaping to ease credit assignment:
\begin{equation}
r_t
=
\lambda_{\Delta}\,\mathrm{clip}(\Delta_t,-0.1,0.5)
+
\lambda_{\mathrm{best}}\,\max(0,\mathrm{IoU}_t-\mathrm{IoU}^*_{t-1})
-
c(a_t)
-
\lambda_{\mathrm{inv}}\cdot \mathbbm{1}[\text{invalid}]
+
r^{\mathrm{pt}}_t,
\end{equation}
where $\mathrm{IoU}^*_{t-1}=\max_{k\le t-1}\mathrm{IoU}_k$ is the best-so-far IoU.
We assign a tool-dependent call cost $c(a_t)=\lambda_{\mathrm{cost}}\cdot \kappa(a_t)$ with multipliers $\kappa(\cdot)$ (e.g., higher for segmentation of 2.5 than for geometric transforms of 1) to discourage excessive tool usage. We adopt $\lambda_{\delta}$ as 1, $\lambda_{best}$ as 0.5, $\lambda_{inv}$ as 1. Then the final $R_{\mathrm{process}}$ is calculated as:
\begin{equation}
R_{\mathrm{process}} = \eta \sum_{t=1}^{T} r_t,
\end{equation}
where we default $\eta$ as 1 during training.

\paragraph{Point-level sparsity/novelty (within $R_{\mathrm{process}}$).}
When invoking the segmentation tool, we additionally encourage \emph{novel} point prompts while penalizing redundant clicks:
\begin{equation}
r^{\mathrm{pt}}_t
=
\mathbbm{1}[\Delta_t>\epsilon]\Big(
\rho\cdot N_{\mathrm{new}}
-
\beta_{\mathrm{pt}}\cdot \mathrm{Redund}
\Big),
\end{equation}
where $N_{\mathrm{new}}$ counts points whose distance to the historical point set is at least $d_{\min}$ of 8, and $\mathrm{Redund}$ accumulates a normalized penalty for points closer than $d_{\min}$, encouraging sparsity and reducing repeated clicks. 
\subsection{Format reward $R_{\mathrm{format}}$.}
We reward syntactically valid, schema-compliant \texttt{<answer>} and \texttt{<tool\_call>} blocks, and penalize unparsable outputs and invalid tool invocations.
Concretely, $R_{\mathrm{format}}$ includes (i) a small bonus for parsable, schema-compliant JSON outputs, and (ii) penalties for invalid tool calls or unparsable generations .

\subsection{Final-answer reward $R_{\mathrm{final}}$.}
Given the final \texttt{<answer>}, we run the segmentation model for each predicted item to obtain masks, and score them against the GT instance set using Hungarian matching.
We define
\begin{equation}
R_{\mathrm{final}}
=
\overline{\mathrm{IoU}}_{\mathrm{match}}
+
0.5\cdot \mathrm{IoU}_{\mathrm{box}},
\end{equation}
where $\overline{\mathrm{IoU}}_{\mathrm{match}}$ is the matched mean IoU over GT connected components, and $\mathrm{IoU}_{\mathrm{box}}$ is the IoU between the union bounding box of predicted masks and the GT union box.
This term directly encourages high quality final masks via both mask IoU and bounding-box IoU.

\paragraph{Episode return.}
We aggregate the above components as Equation.~\ref{eq:Rtotal}
where $\beta$ controls the strength of dense process shaping relative to the final outcome. We adopt $\alpha$ of 1, $\beta$ of 0.5, $\gamma$ of 0.2 as our training parameter.

\section{Baseline Methods.}
\label{append:baseline}
We compare RSAgent with four families of representative methods on both RES and ReasonSeg.
(\textit{1) Discriminative vision--language segmentors without LLM-based controllers.}
This group covers classical RES/RIS networks and open-vocabulary segmentors that rely on task-specific cross-modal encoders or CLIP-style backbones, but do not use a generative LLM/VLM as the decision making controller(e.g., LAVT \cite{yang2022lavt}, ReLA \cite{ReLA},GLEE \cite{glee2024}, DETRIS \cite{huang2025detris}, UniLSeg \cite{unilseg2024liu} and EVF-SAM \cite{EVF_SAM_2024_Zhang} on RefCOCO-style benchmarks, as well as SEEM \cite{seem2023zou}, Grounded-SAM \cite{ren2024groundedsam} and OVSeg \cite{Liang_2023_MCLIP} on ReasonSeg-style evaluations).
(\textit{2) Large (multi-)modal models for single-pass segmentation without explicit reasoning traces.}
These methods leverage large multimodal backbones and predict masks in a mostly one-shot manner, without exposing step-by-step textual reasoning during inference. Representative examples include GLaMM \cite{glamm2024}, GSVA \cite{gsva2024}, SAM4MLLM \cite{sam4mllm2024},  PixelLM \cite{pixellm2024}, and RICE \cite{xie2025rice} on RES, together with early MLLM-based reasoning-segmentation baselines such as LISA \cite{lisa_2024_Lai}, Hyperseg \cite{hyperseg2024}, X-SAM \cite{wang2025xsam}.
(\textit{3) Explicit chain-of-thought (CoT) or reasoning-chain guided segmentation.}
This family explicitly produces multi-step textual explanations or structured reasoning traces (often followed by localization/mask prompting) before committing to a mask. Typical examples include RSVP \cite{rsvp}, as well as RL-enhanced reasoning-chain approaches such as Seg-Zero \cite{segzero}, Seg-R1 \cite{segr1} and SAM-R1 \cite{samr1}.
(\textit{4) Multi-round tool-calling agents with iterative refinement.}
Different from one-shot prediction or single-pass prompt generation, these agent-style frameworks iteratively call external segmentation tools, inspect intermediate masks, and update prompts over multiple rounds. A representative example is SAM3 Agent \cite{carion2025sam3}, which uses an MLLM to repeatedly propose prompts for SAM3 and refine them based on tool feedback.
Together, these four groups cover non-LLM discriminative architectures, one-shot MLLM segmentors, explicit CoT-based reasoning segmentation, and multi-round tool-using agents, providing a comprehensive comparison ground for RSAgent.

%%%%%%%%%%%%%%%%%%%%投稿的额外附录内容%%%%%%%%%%%%%%%%%%%%%%
% \paragraph{Analysis of our generated dataset.}
% As shown in Table.~\ref{}, our $D_\mathrl{SFT}$ completely follow the data generation pipeline, consisting of 5K reasoning trajectories.

% \section{Introduction of our Training prompts.}

% \section{Training Details.}

% \section{More Visualization Examples.}

% \section{Limitation and Future Work.}

%%%%%%%%%%%%%%%%%%%%%%%%%%%%%%%%%%%%%%%%%%%%%%%%%%%%%%%%%%%%%%%%%%%%%%%%%%%%%%%
%%%%%%%%%%%%%%%%%%%%%%%%%%%%%%%%%%%%%%%%%%%%%%%%%%%%%%%%%%%%%%%%%%%%%%%%%%%%%%%

\end{document}